\documentclass[preprint,12pt,3p]{elsarticle}

\usepackage{amssymb}

\usepackage{graphicx}
\usepackage{times}
\usepackage{graphicx}
\usepackage{amsmath}
\usepackage{bm}
\usepackage{color}
\usepackage[table]{xcolor}
\usepackage{booktabs}
\usepackage{multirow}
\usepackage{varwidth}
\usepackage{caption} 
\usepackage[numbers]{natbib}
\DeclareMathOperator*{\argmax}{\arg\!\max}

\newcommand*{\rom}[1]{\expandafter\@slowromancap\romannumeral #1@}
\long\def\IGNORE#1{} \long\def\COMMENT#1{}
\usepackage[pagebackref=true,breaklinks=true,letterpaper=true,colorlinks,bookmarks=false]{hyperref}

\journal{JVCI}

\begin{document}

\begin{frontmatter}


\title{Discovering Spatio-Temporal Action Tubes}

\author[label1]{Yuancheng Ye}
\address[label1]{The City College and Graduate Center, City University of New York}

\ead{yye@gradcenter.cuny.edu}

\author[label5]{Xiaodong Yang}
\address[label5]{NVIDIA Research}
\ead{xiaodongy@nvidia.com}

\author[label1]{YingLi Tian\corref{cor1}}
\ead{ytian@ccny.cuny.edu}

\cortext[cor1]{Corresponding author}

\begin{abstract}   
In this paper, we address the challenging problem of spatial and temporal action detection in videos. We first develop an effective approach to localize frame-level action regions through integrating static and kinematic information by the early- and late-fusion detection scheme. With the intention of exploring important temporal connections among the detected action regions, we propose a tracking-by-point-matching algorithm to stitch the discrete action regions into a continuous spatio-temporal action tube. Recurrent 3D convolutional neural network is used to predict action categories and determine temporal boundaries of the generated tubes. We then introduce an action footprint map to refine the candidate tubes based on the action-specific spatial characteristics preserved in the convolutional layers of R3DCNN. In the extensive experiments, our method achieves superior detection results on the three public benchmark datasets: UCFSports, J-HMDB and UCF101.  
\end{abstract}

\begin{keyword}
Spatio-temporal action detection, deep neural networks
\end{keyword}

\end{frontmatter}

\section{Introduction}
Action recognition in videos embodies either two primary tasks, i.e., action classification and action detection. Most prior studies \cite{xiaodong2016multilayer, karpathy2014large, yang2014eccv, donahue2015long, snvpami, pichao-cviu} focus on the task of action classification, which assigns an action label to the whole video. By contrast, action detection not only identifies the category of an action but also localizes where the action happens in a video. Although a number of methods have been successfully proposed for action classification, action detection in wild videos still remains as a challenging task and receives far less attention.

Most methods developed for action detection consider either only localizing actions in spatial \cite{delaitre2011learning, yu2015fast} or only detecting temporal boundaries of actions \cite{budget-aware, r-c3d, language-model}, which are inadequate for some more advanced video analytic applications such as video segmentation and events detection. Some algorithms recently proposed in \cite{Escorcia2018, gkioxari2015finding, Hou2017, Kalogeiton2017, MettesECCV2016, singh2017, van2015apt, weinzaepfel2015learning, zhuiccv2017} make a further step to simultaneously detect actions in both spatial and temporal domains. These methods typically share two principle ingredients: (1) detecting action regions on each individual frame and (2) linking the detected action regions throughout the whole video sequence.

\begin{figure}[t]
\centering
\includegraphics[width=\linewidth]{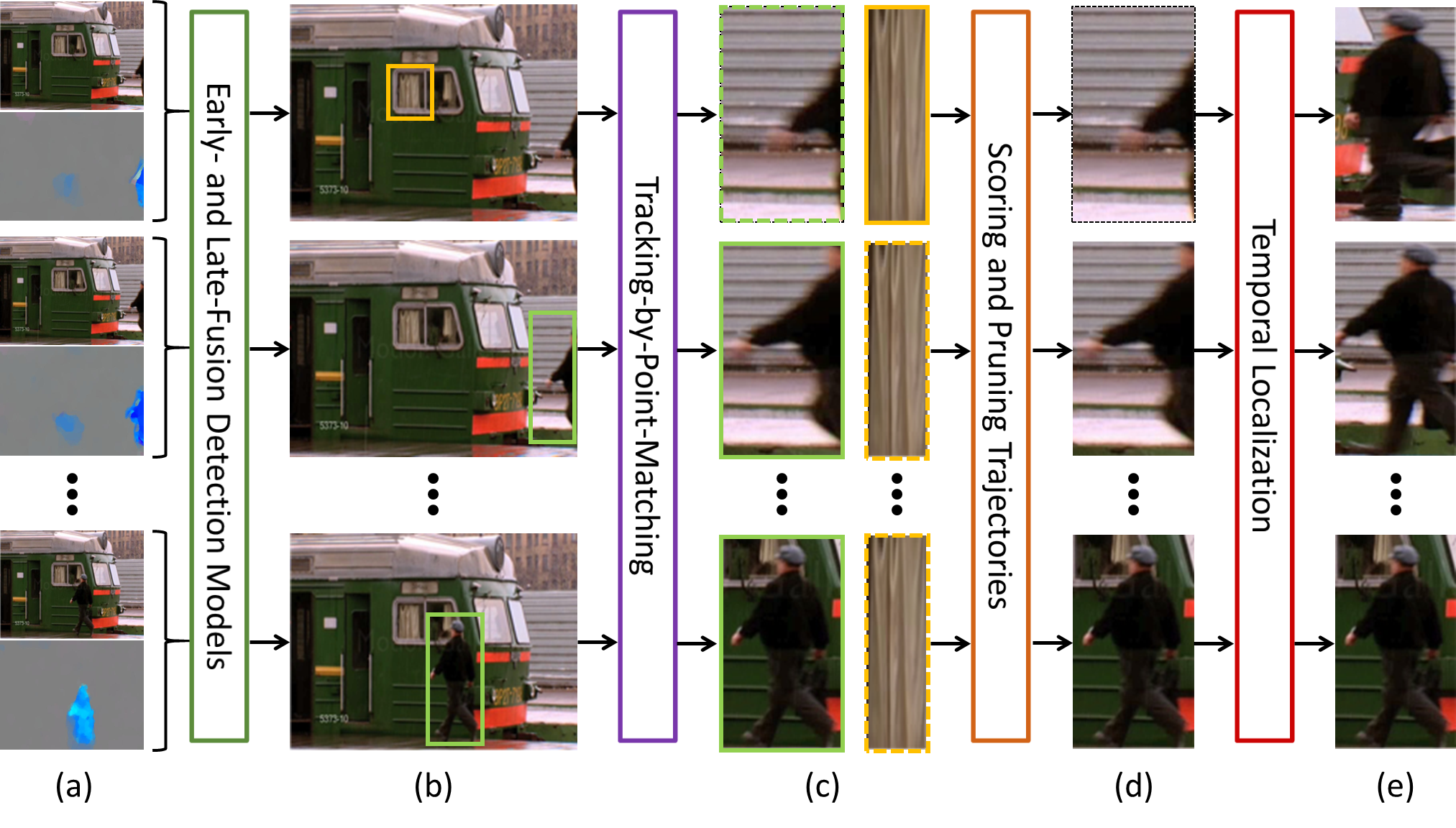}
\caption{An overview of the proposed action detection framework. (a): continuous static images and corresponding colorized optical flow images in a video. (b): detected action regions on individual frames. (c): multiple candidate tubes produced by the tracking-by-point-matching algorithm, where the regions enclosed by solid lines are the detected action regions in (b), and the ones enclosed by dash lines are predicted in the tracking process. (d): the false tube, i.e., the one enclosed by yellow lines in (c), is removed after the scoring and pruning procedures. (e): the final action tube with refined temporal boundary.}
\label{fig:pipeline}
\end{figure}

In this paper, we proceed along with this direction to perform spatial and temporal action detection. Specifically, we develop an effective frame-level action region detection approach by the proposed early- and late-fusion of static appearances and dynamic motions. In the early-fusion, we stack the image channels of both static and optical flow images as an unified input, while the late-fusion exploits two-stream networks based on static and optical flow images separately and combines their individual outputs. Results of both early- and late-fusion models are integrated as our detected action regions at the frame-level. In order to establish the important temporal connections among the detected action regions, we propose a tracking-by-point-matching algorithm to stitch the discrete action regions into continuous action tubes by leveraging on the robust region proposals and accurate point matching. To model the short-term motion cues and long-term temporal context, we harness the recurrent 3D convolutional neural network (R3DCNN) to classify action categories and determine temporal extents of the action tubes. Additionally, we introduce an action footprint map to prune the candidate tubes by taking advantage of the discriminative spatial attention retained in the convolutional layers of R3DCNN. We refer a detected spatio-temporal action sequence as an action tube. Figure~\ref{fig:pipeline} illustrates the pipeline of the proposed framework to generate spato-temporal action tubes.

The main contributions of this paper are three folds. Firstly, the early- and late-fusion models are integrated to effectively fuse static appearances and dynamic motions to detect action regions. Secondly, we propose a new tracking-by-point-matching algorithm to connect detected regions in videos. At last, the action footprint map is utilized to further prune false tubes. As shown in Table~\ref{table:results}, our method achieves superior results in comparison to the competing algorithms. A variety of ablation studies are conducted in the experiments for the purpose of in-depth analysis of each component in our approach.

The rest of the paper is organized as follows. Section 2 presents a brief review of the related work on action recognition. Section 3 introduces the early- and late-fusion schemes for action region detection at frame-level. In Section 4, we provide  the detailed procedures of the tracking-by-point-matching algorithm to transform the detected action regions to candidate tubes. Section 5 describes the refining procedures to produce final action tubes. Experimental results are presented in Section 6. Finally, we summarize the remarks of this paper in Section 7.


\section{Related Work}
As one of the primary research directions in action recognition, action classification has drawn far more attention than action detection. Driven by the success of deep learning in image classification \cite{krizhevsky2012imagenet}, many studies have explored to use convolutional neural networks (CNNs) for video classification \cite{donahue2015long, karpathy2014large}. In addition to applying the 2D convolutions to individual video frames, Tran et al. \cite{tran2015learning} introduced the C3D model to simultaneously learn spatial and temporal features using the 3D convolutions with a buffer of video frames. Xie et al. \cite{xie2017rethink} studied various 3D convolutional networks for video understanding tasks to be more accurate and efficient. Feichtenhofer et al. \cite{Feich2016} proposed an effective network architecture for spatio-temporal fusion of video snippets and studied different ways of fusing appearance and motion information. PreRNN was proposed in \cite{prernn} to transform convolutional networks to recurrent networks for various video understanding tasks including action classification. 

For action detection in videos, most research focuses on either spatial or temporal detection. To localize actions in spatial, Yu and Yuan \cite{yu2015fast} integrated the actionness score with a greedy method to generate action proposals on individual frames. Jain et al. \cite{jain2014action} computed spatio-temporal bounding boxes by merging a hierarchy of supervoxels and classified the candidate boxes by motion features. Tian et al. \cite{tian2013spatiotemporal} extended the deformable part models to videos for spatial action detection. To determine the temporal boundaries of actions, 
a sliding window approach was introduced in \cite{pyramid-score} to build pyramid representations in order to capture motion information at multiple resolutions. Oneata et al. \cite{oneata2014efficient} proposed to replace the sliding window approach with a more efficient branch-and-bound search. 
Escorcia et al. \cite{escorcia} introduced a temporal segment proposal algorithm based on C3D and LSTM. R-C3D was proposed in \cite{r-c3d} to save computational costs by sharing convolutional features between proposal and classification stages. A recurrent policy network was recently developed to perform temporal action detection within a time budget \cite{budget-aware}.

Gkioxari and Malik \cite{gkioxari2015finding} proposed to detect spatial regions on individual frames and then link them according to spatial overlapping and classification scores. However, action consistency is not considered in the linking process which might result in inferior performance if there are multiple targeting actions in a video. Weinzaepfel et al. \cite{weinzaepfel2015learning} employed a standard tracking algorithm to track the interest regions over frames to produce a bunch of tubes. The tracking procedure in \cite{weinzaepfel2015learning} depends on the constrained neighboring windows of the detected regions and is therefore difficult to deal with the large displacements caused by human movements, a common difficulty for action detection in wild videos. Additionally, the tracking error can be propagated and accumulated into the following tracked regions, which would adversely impact the overall performance. Peng and Schmid \cite{peng2016multi} embedded a multi-region scheme in the Faster R-CNN model, which provides complementary information on body parts to refine the action locations, and adopted a linking method to connect the frame-level detections. 
Mettes et al. \cite{MettesECCV2016} attempted for weakly-supervised action detection using only points on a sparse subset of frames instead of action boxes. Kalogeiton et al. \cite{Kalogeiton2017}  employed the SSD framework \cite{LiuECCV2016} and proposed the action tubelet detector that takes as input a sequence of frames and outputs tubelets. Hou et al. \cite{Hou2017} proposed a unified tube convolutional neural network to recognize and localize action based on 3D-CNN. Singh et al. \cite{singh2017} presented a deep-learning framework based on SSD with an efficient online
algorithm to incrementally construct and label ‘action
tubes’ from the SSD frame level detections for real-time multiple spatio-temporal  action localization and classification. Zhu et al. \cite{zhuiccv2017} proposed a spatio-temporal convolutional network which consists of a temporal convolutional regression network and a spatial regression network by empowering convolutional LSTM with regression capability. Escorcia et al. \cite{Escorcia2018} developed an actor-supervised architecture that exploits the inherent compositionality of actions in terms of actor transformations to localize actions. Gu et al. \cite{gu2018} introduced a new dataset for human action detection called atomic visual actions (AVA) and proposed an approach for action localization that builds upon the current state-of-the-art methods. 


\section{Action Region Detection}
\begin{figure*}[t]
\centering
\includegraphics[width=0.8\linewidth]{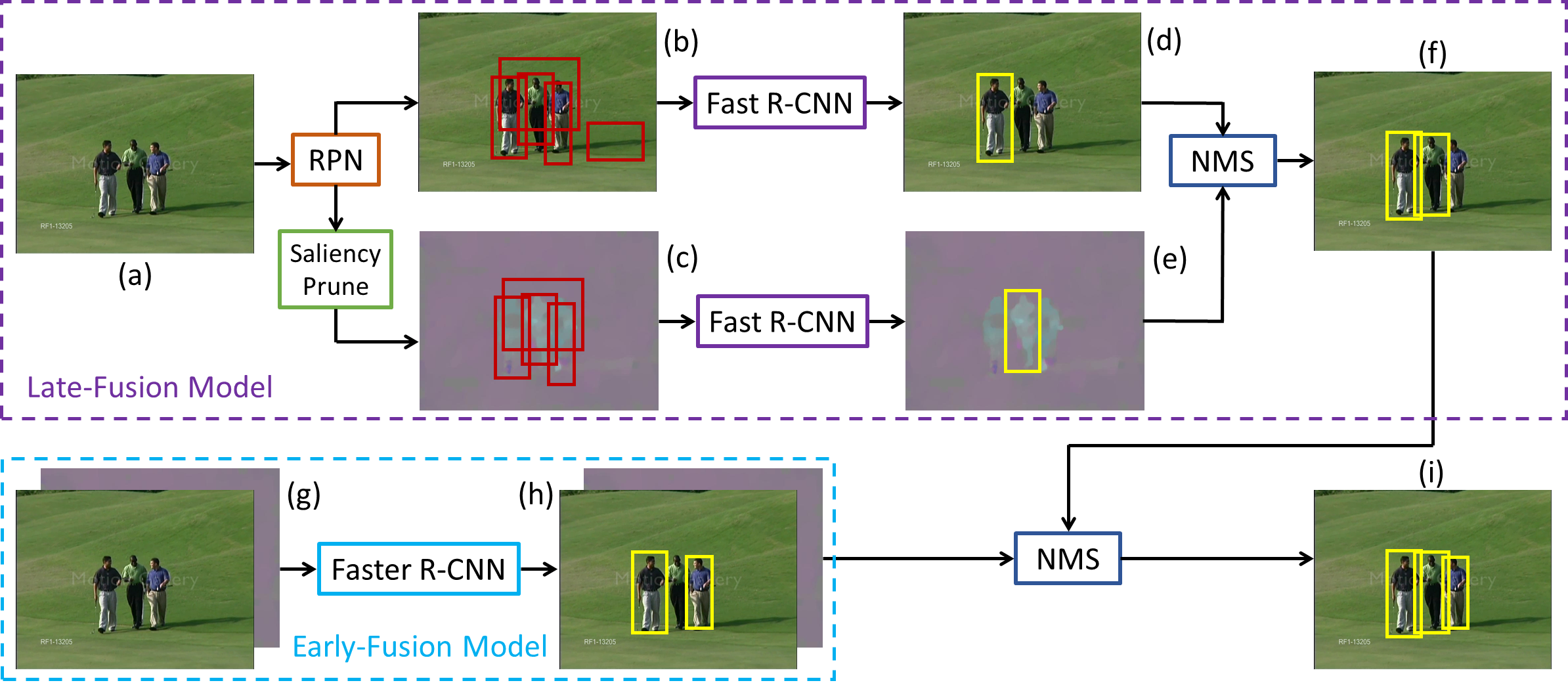}
\caption{An overview of the early- and late-fusion models for action region detection at the frame-level. (a): the input frame to the late-fusion model. (b): region proposals produced by RPN. (c): region proposals pruned by motion saliency. (d): action regions detected in the static stream. (e): action regions detected in the dynamic stream. (f): detection results by the late-fusion. (g): the composite input to the early-fusion model by concatenating the static and optical flow images. (h): detection results by the early-fusion. (i): the final action region detection by integrating the early- and late-fusion models.}
\label{fig:two-stream}
\end{figure*}

In the previous studies, R-CNN \cite{Girshick14} is widely used to detect spatial regions on individual frames. Features are typically extracted from appearance and motion models separately and concatenated as input to a linear SVM classifier. In our framework, we employ Faster R-CNN \cite{ren2015faster} as the backbone network for action region detection. Faster R-CNN consists of region proposal network (RPN) and Fast R-CNN \cite{girshick2015fast}. RPN generates region proposals and Fast R-CNN determines their categories. RPN and Fast R-CNN share the same weights in the convolutional layers. In contrast to the generic object region detection, temporal context and motion cues play a critical role in detecting action regions. Therefore to better exploit the advantages of Faster R-CNN and integrate the appearance and motion information, we develop two complementary modules including the early- and late-fusion models to detect the frame-level action regions by effectively fusing the static and kinematic features, as illustrated in Figure~\ref{fig:two-stream}.

In the early-fusion model, we concatenate the RGB channels of both static and colorized optical flow images as a composite input to Faster R-CNN. Accordingly we modify the filters of the first convolutional layer of Faster R-CNN to make the pre-trained weights compatible to the 6-channel input image, i.e., replicating the filter weights along the depth dimension and dividing them by $2$ to compensate for the numerical scaling change. The rest layers of Faster R-CNN remain the same. By early fusing the static and optical flow images as one input, we enforce Faster R-CNN to jointly learn the inter-related clues between appearances and motions for detecting action regions.

In late-fusion we adopt two-stream Faster R-CNNs to process static and dynamic information separately and then integrate the results by non-maximum suppression (NMS). For the static stream, both RPN and Fast R-CNN are trained on the static images. The dynamic stream shares the RPN from the static stream to yield action region proposals. Since a number of action regions rarely contain any motions on the optical flow images, we apply a simple saliency pruning to remove such motionless regions. Fast R-CNN for the dynamic stream is then separately trained on the optical flow images. In the end, we integrate the detection results of the two fusion modules as in Figure~\ref{fig:two-stream}. Our evaluations in Table~\ref{table:fusion-impact} demonstrate that combining the two fusion models is effective to improve the overall performance compared to any individual fusion module. This suggests that merging the appearance and motion information at different stages through Faster R-CNN is able to provide complementary and mutually corrected action regions.

\begin{figure*}[t]
\centering
\includegraphics[height=14cm, width = 12cm]{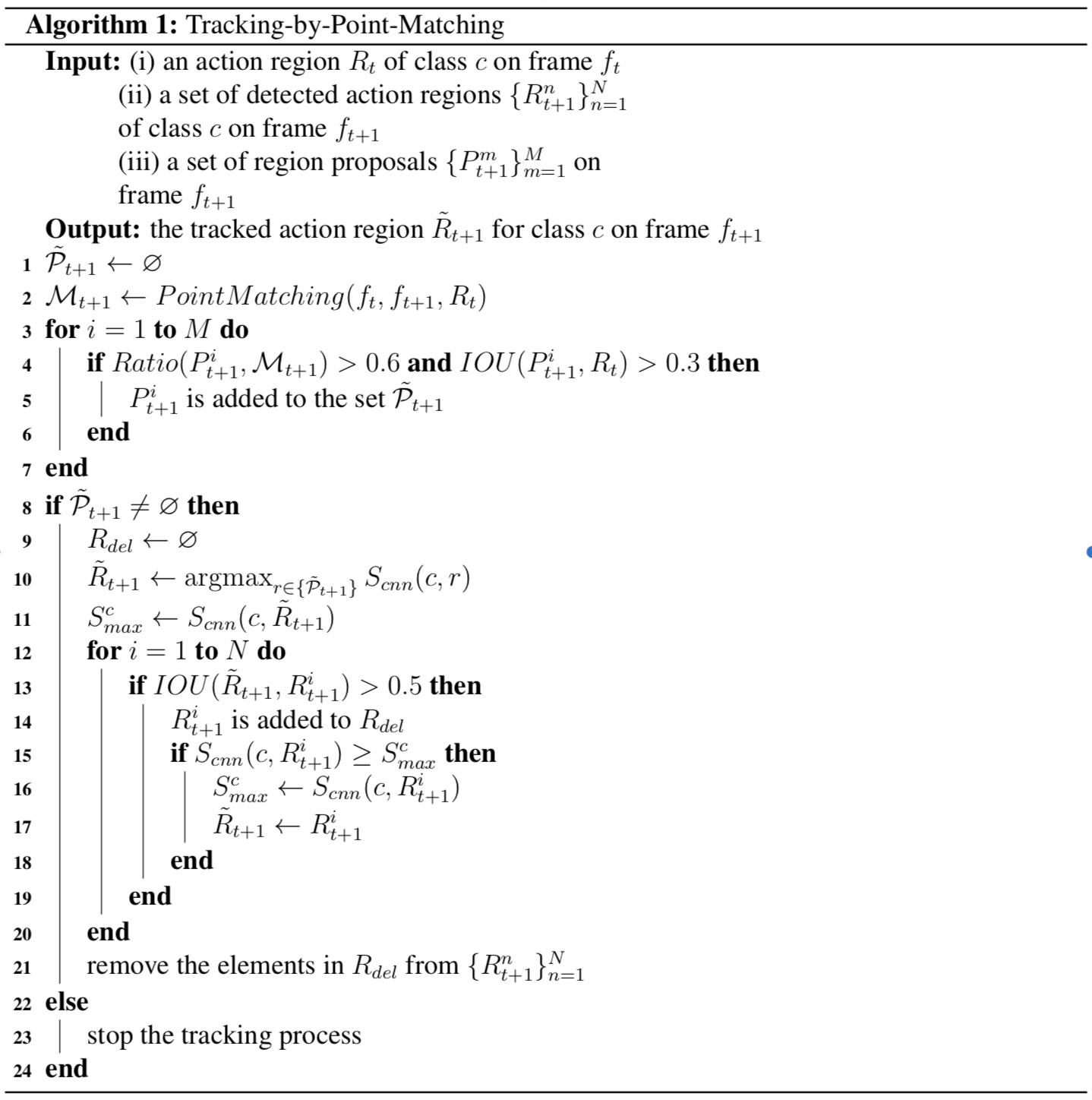}
\caption{Algorithm of the tracking-by-point-matching to connect detected action regions.}
\label{fig:algorithm}
\end{figure*}

\section{Tracking-by-Point-Matching}

The detected action regions serve as the building blocks to construct spatio-temporal action tubes. These regions however can be discontinuous in temporal, moreover the action region detector may assign different labels to the same action of different frames. To inject the temporal connections into the detected action regions across a whole video, most previous methods employ either linking or tracking-by-detection algorithms. The linking method in \cite{gkioxari2015finding} connected the detected action regions by maximizing their combination scores, while the tracking-by-detection algorithm in \cite{weinzaepfel2015learning} established action tracks through selecting regions with the highest scores in the confined neighborhood of detected action regions. Nevertheless, the linking method could fail if there are multiple targeting actions in one video, and the tracking-by-detection algorithm performs not great when an action exhibits large motion displacements. 

In order to overcome these restrictions, we propose a new tracking-by-point-matching algorithm to impose temporal and semantic continuity on the action sequence. Unlike the traditional tracking paradigms that confine the tracking procedures with a manually defined neighboring space, our algorithm can handle large motion displacements by leveraging on the high recall of action region proposals and accurate point matching. 

Algorithm 1 shows the tracking-by-point-matching algorithm between two consecutive frames in the forward direction. Let $\{R_t^n\}_{n=1}^{N}$ denote a set of action regions detected on frame $f_t$, and $\{P_t^m\}_{m=1}^{M}$ is a set of action region proposals output by RPN of the action region detection model on the same frame. $PointMatching(f_t, f_{t+1}, R_t^n)$ is a point matching function which takes two consecutive frames and an action region as input, and outputs a set of matched points $\mathcal{M}_{t+1}$ on frame $f_{t+1}$ corresponding to the points from the action region $R_t^n$. $Ratio(P_{t+1}^m, \mathcal{M}_{t+1})$ computes the proportion of points of $\mathcal{M}_{t+1}$ contained in the region proposal $P_{t+1}^m$. To ensure the predicted region has enough contextual similarity and spatial continuity with the action region on the previous frame, only the proposals with sufficient matched points and overlapped area with the previous action region are added into the region candidate pool $\tilde{\mathcal{P}}_{t + 1}$.

If there is no more tracked regions or $\tilde{\mathcal{P}}_{t + 1}$ is empty, we assume that the targeting action vanishes in the following frames and the tracking procedure should be terminated. Otherwise, the action is not finished and the action region should be propagated onto the next frame. $S_{cnn}(c, r)$ calculates the score of a region $r$ to class $c$ based on the action region detection model. The region with the highest score in the candidate pool $\tilde{\mathcal{P}}_{t + 1}$ is denoted as $\tilde{R}_{t + 1}$ which is used to replace the action region $R_{t + 1}^i$ if they are sufficiently overlapped, and $R_{t + 1}^i$ is removed from the pool of untracked action regions. This process not only avoids redundant computations but also prevents error propagation in the following tracking procedures. Finally, the region $\tilde{R}_{t + 1}$ is output as the tracked action region on frame $f_{t+1}$. We empirically set the related thresholds in our algorithm. By applying the tracking-by-point-matching algorithm on each detected action region recursively in both forward and backward directions, we can generate multiple candidate tubes throughout a video. 


\section{Refinement of Candidate Tubes}
We can roughly acquire the spatial and temporal locations of an action after the tracking-by-point-matching process. However, if an action region with a certain class is falsely detected by the frame-level detection model, the resulting candidate tubes from this action region are incorrect as well. In this section we focus on how to prune the plausible candidate tubes and hone their temporal boundaries to generate the final action tubes.

\subsection{Architecture of R3DCNN}
We first measure the credibility of each candidate tube by two sources: 1) the average score of action regions in the tube computed by the frame-level detection model and 2) the average score of short clips in the tube computed by R3DCNN. We employ R3DCNN to explicitly take into account the local spatio-temporal cues and the global temporal evolution in a candidate tube. The architecture of R3DCNN consists of the pre-trained C3D on Sports1M \cite{tran2015learning} for short-term spatio-temporal feature extraction and the PreRNN structure \cite{prernn} for long-term temporal modeling.

RNN is a sequence-based network to model the temporal progress through a hidden state $\bm{h}_t$ at time step $t$, and its activations also dependent on that of the previous time step:
\begin{equation}
\bm{h}_t = \sigma(\bm{W}_{ih}\bm{y}_t + \bm{W}_{hh}\bm{h}_{t-1} + \bm{b}_h),
\end{equation}
where $\sigma$ is the activation function, $\bm{W}_{ih}$ is the input-to-hidden weight matrix, $\bm{W}_{hh}$ is the hidden-to-hidden weight matrix, $\bm{y}_t$ is the input feature, and $\bm{b}_h$ is the bias. In most vision tasks RNN is built on CNNs that are pre-trained on large-scale datasets for better generalization. However in the traditional RNN, both $\bm{W}_{ih}$ and $\bm{W}_{hh}$ are randomly initialized. It therefore requires to train such a recurrent layer from scratch even if a pre-trained CNN is used for feature extraction (e.g., the pre-trained C3D in our case). We adopt the recently proposed PreRNN \cite{prernn} to fuse the recurrent layer with the fully connected layer of C3D to preserve the important generalization property. 

Suppose the output of a fully connected layer of C3D at time step $t$ is:
\begin{equation}
\bm{y}_t = \sigma(\bm{W}_{io}\bm{x}_t + \bm{b}_y),
\end{equation}
where $\bm{W}_{io}$ is the pre-trained input-to-output weight matrix, $\bm{x}_t$ is the output of previous feed-forward layer, and $\bm{b}_y$ is the bias. PreRNN transfers it into a recurrent layer through:
\begin{equation}
\bm{y}_t = \sigma(\bm{W}_{io}\bm{x}_t + \bm{W}_{hh}\bm{y}_{t-1}+ \bm{b}_y).
\end{equation}
This recurrent structure, initialized by the fully connected layer of C3D, only introduces a single hidden-to-hidden weight matrix $\bm{W}_{hh}$ that needs to train from scratch, while other weight matrices have already been pre-trained and can be just fine-tuned. We choose to use PreRNN to model the temporal connections because of its simple structure design and superior or on par performance to other complex variants of recurrent networks as shown in Table~\ref{table:score-impact}.

\subsection{Scoring and Pruning Action Tubes}
In order to take into account the local region information and long-term temporal context, we apply both action region detection model and R3DCNN to each candidate tube to determine the final score to a class $c$:
\begin{equation}
S_{traj}^c = S_{avg-cnn}^c + S_{avg-rnn}^c,
\end{equation}
where $S_{avg-cnn}^c$ is the average score of action regions in the tube by the frame-level detection model and $S_{avg-rnn}^c$ is the average score of sliced clips in the tube by R3DCNN. Apart from the addition of the two scores, multiplicative operation \cite{pichao-mm} can be applied as well. The action label  $l$ of a candidate tube is $l = \argmax_i S_{traj}^i$, and the final score is $S_{traj}^l$.

We observe two types of false tubes---overlapped and drifted tubes---to remove from the candidate tubes. The overlapped tubes are mostly aroused from ambiguous classification of the frame-level action region detection, e.g., the detection model might detect two action regions $r_{t_1}^{c_1}$ and $r_{t_2}^{c_2}$ for the same targeting action. After the tracking-by-point-matching algorithm, two tubes with different action classes ($c_1$ and $c_2$) can be produced at the same action area in the video. The drifted tubes are not unusual for the videos with very complex background where the falsely matched points can be generated from cluttered objects and people.

It is straightforward to remove the overlapped tubes in our framework. If the spatio-temporal IOU of two tubes is greater than a threshold ($0.3$ in our experiments), the candidate tube with lower action tube score $S_{traj}^l$ is excluded. We define the spatio-temporal IOU between two candidate tubes as the product of the temporal IOU and the average of spatial IOU over all overlapped frames.

In order to prune the drifted tubes, we propose an action footprint map based on the convolutional layers of R3DCNN to leverage the preserved action-specific spatial characteristics. Suppose $s$ is the spatial size of feature maps of a convolutional layer in R3DCNN, and $d$ is the number of feature maps. For a candidate tube, a sequence of feature maps can be extracted $\mathcal{F} = \{\bm{f}_t; t = 1,\ldots,T\}$, where $T$ is the number of sliced clips in the tube, and $\bm{f}_t \in \mathbb{R}^{s \times s \times d}$ represents the feature maps computed at the selected convolutional layer of the $t$-th clip. We convert $\bm{f}_t$ into $s \times s$ features each of which is a $d-$dimensional descriptor so that each candidate tube generates $s \times s \times T$ feature descriptors $\bm{z}_i \in \mathbb{R}^d$. We aggregate these feature descriptors on a set of pre-defined spatial neighboring cells over the selected convolutional layer. Let $C$ indicate the pre-defined cells and $C_j$ denote the $j$-th cell. A spatial cell on the selected convolutional layer is then represented by: 
\begin{equation}
\bm{c}_j = \mathcal{H}(\{\bm{z}_i\}_{i \in C_j}), j = 1,\ldots,|C|,
\end{equation}
where $\mathcal{H}$ is the Fisher vector coding operator \cite{perronnin2007fisher} that aggregates $\bm{z}_i$ within a local spatial region across the whole action tube. We make use of $\bm{c}_j$ to perform action classification and the accuracy $\alpha_j^l$ associated with the spatial cell $C_j$ (and the corresponding receptive field on video frames) signifies how discriminative this local spatial region is in the candidate tube for classifying this action. We in the end transfer the classification accuracy $\alpha_j^l$ to an action footprint factor $w_j^l$ by the softmax function: $w_j^l = \exp(\alpha_j^l)/\sum_{k=1}^{|C|}\exp(\alpha_k^l)$.

\begin{figure}[t]
\centering
\includegraphics[width=0.8\linewidth]{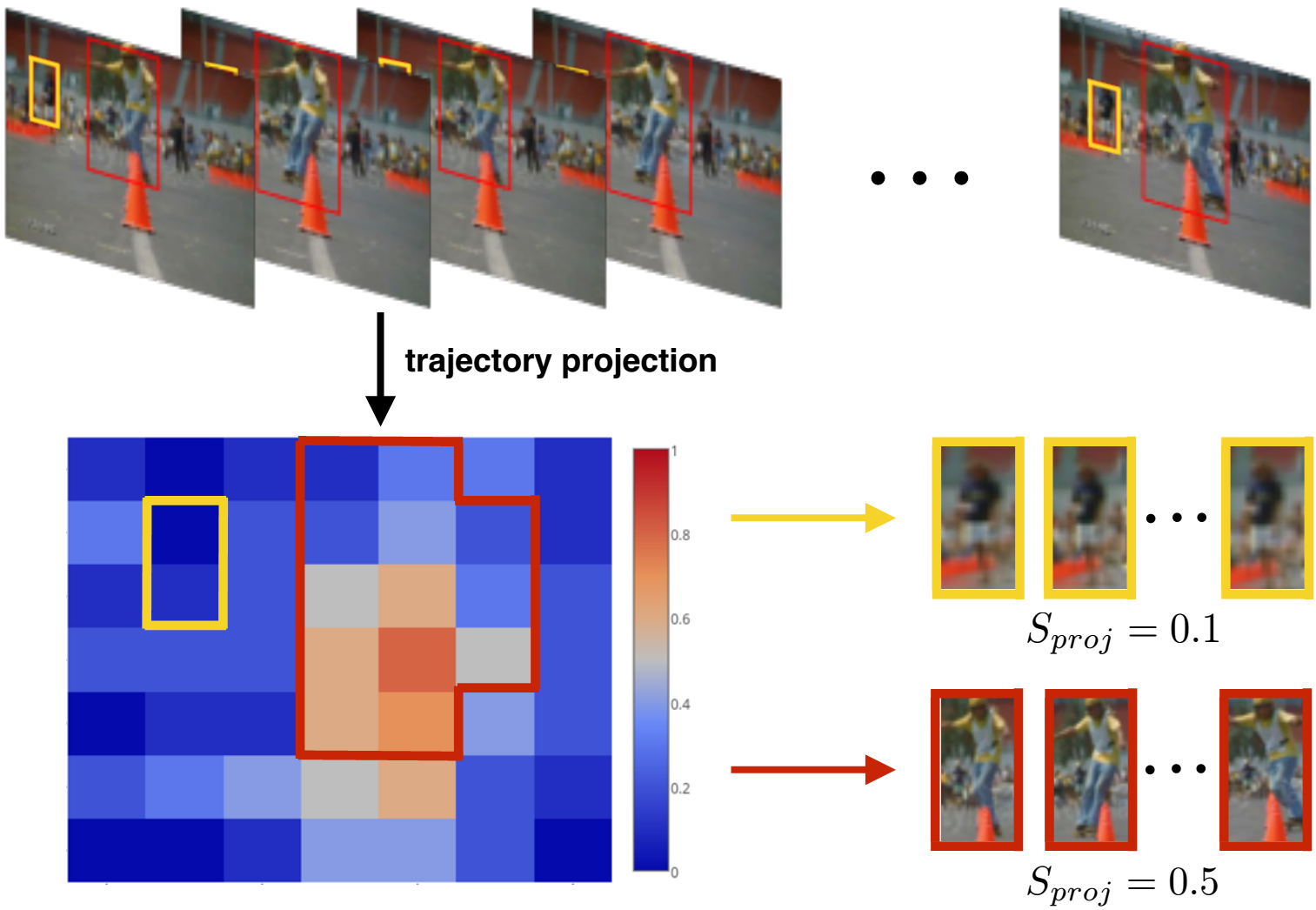}
\caption{Illustration of the action footprint map. The enclosed cells by yellow and red lines on the map are the projected areas of candidate tubes. The off-focusing action tube surrounded by yellow lines with a lower average footprint factor is removed.}
\label{fig:footprint}
\end{figure}

As demonstrated in Figure~\ref{fig:footprint}, we can compute an action footprint map at a selected convolutional layer of R3DCNN for each action category. Since $w_j^l$ represents the spatial footprint in the corresponding receptive field of action $l$ or how discriminative the spatial cell $C_j$ is for classifying action $l$, we can take advantage of this capability to prune the drifted action tubes. We project a candidate tube of class $l$ to its action footprint map and get the projected area with a set of overlapped cells $O$. We denote the average footprint factor in the projected area as $S_{proj} = \sum_{i \in O} w_i^l / |O|$, and the average footprint factor over the entire action footprint map as $S_{map} = \sum_{i \in C} w_i^l / |C|$. We argue that if $S_{proj}$ of an action tube is less than $S_{map}$, this tube is off the spatial focusing area for its action category and therefore should be removed. Our ablation study in Table~\ref{table:score-impact} shows that the action footprint map is effective in pruning the drifted tubes and improving the overall performance. Moreover, the features used to compute the maps come for free because they have been already extracted during the forward pass of R3DCNN.

\subsection{Temporal Localization}
To decide the temporal boundaries, most previous methods employ a set of temporal sliding windows of different sizes on the extracted tracks with varied steps to localize actions.

Instead of using this computationally expensive approach, our temporal localization is naturally leveraged on the temporal propagating characteristic of R3DCNN. Suppose that a candidate tube with action class $l$ is sliced into $K$ clips, and the scores of these clips are $\big \langle S_1,\ldots,S_{K}\big \rangle$. If $S_i$ and $S_j$ are the first and last entries in this score sequence that are lower than a threshold ($0.3$ in our experiments), the action tube is then temporally localized between the $i$-th and $j$-th clips. The elements between $S_i$ and $S_j$ are not considered because each action tube generated by the tracking-by-point-matching algorithm contains only a single action that spans for a continuous period.

\section{Experiments}
In this section, we extensively evaluate our proposed method on the three benchmark datasets for spatial and temporal action detection: UCFSports \cite{rodriguez2008action}, J-HMDB \cite{jhuang2013towards}, and UCF101 \cite{soomro2012ucf101}. Experimental results show that our algorithm achieves superior results on the three datasets. A variety of ablation studies are conducted to analyze the impact of each component in our approach. 

\subsection{Datasets}
UCFSports \cite{rodriguez2008action} contains 150 videos of 10 action classes with annotated bounding boxes available on each frame. In our experiments we follow the standard experimental setting as defined in \cite{lan2011discriminative}. J-HMDB \cite{jhuang2013towards} consists of 928 videos of 21 action classes. We use the ground truth provided by \cite{gkioxari2015finding} and report the average results over the three standard training and testing splits in our experiments. UCF101 \cite{soomro2012ucf101} is originally dedicated to action classification with more than 13,000 videos and 101 classes. For a subset of 24 classes and 3,207 videos, the spatio-temporal extents of the actions are annotated. As being consistent with the previous studies \cite{peng2016multi, weinzaepfel2015learning}, we report the performance on the first split of three training and testing splits of this dataset. We use the mean average precision (mAP) as our evaluation metric for spatial and temporal action detection. The comparison of our approach with other methods is also reported by the metric of area under the curve (AUC) on both UCFSports and J-HMDB datasets.

\subsection{Implementation Details}
We implement the early- and late-fusion models for action region detection in Caffe \cite{jia2014caffe} and compute optical flow by EpicFlow \cite{revaud2015epicflow}. The Faster R-CNN model is pre-trained on the object detection dataset of PASCAL VOC 2012 and fine-tuned on each action detection dataset.

In the tracking-by-point-matching algorithm, we use deep matching algorithm \cite{weinzaepfel} to match points on the static frames only and ignore the optical flow images. For each video, a dynamic pool of untracked regions is constructed from the detected action regions on all frames, and the tracking procedure terminates when the dynamic pool is empty.

The input to R3DCNN is a sequence of 16-frame clips. The C3D network is pre-trained on the Sports1M dataset \cite{karpathy2014large} and fine-tuned together with PreRNN on the ground truth action tubes. Considering the trade-off between spatial resolution and feature representation capacity, we select \texttt{conv4} in R3DCNN as the convolutional layer to compute the action footprint map and use $2 \times 2$ as the size of a local spatial cell to aggregate the feature descriptors, consequently the action footprint map is of size $7 \times 7$ as shown in Figure~\ref{fig:footprint}.

\subsubsection{Evaluation metrics} If a detected action region has an IOU value with any ground-truth bounding boxes or tubes larger than a threshold $\sigma$ and the assigned action label equals to that of the ground-truth, then this detection is considered as correct. Moreover, the IOU value between two action tubes is defined as the multiplication of the temporal IOU value and the average of the spatial IOU values over all overlapping frames. Our results are reported on all three datasets, measured by the mean average precision (mAP). Both frame level and video level mAPs are reported.  The area under the curve (AUC), which measures the area under the ROC curve, is also reported and compared with other state-of-the-art results.

\subsection{Ablation Studies}
In this section, we focus on the UCFSports and J-HMDB datasets to inspect and understand the impact of each component and present the details of different types of false detections in our method.

\begin{table}
\centering
\resizebox{\linewidth}{!}{
\begin{tabular}{ l c c c c c }
 \hline
  \multirow{2}{*}{Methods} & \multicolumn{2}{c}{Recall-Track} & \hspace{15pt} & \multicolumn{2}{c}{Video-mAP}\\ 
 \cline{2-3} \cline{5-6}
  & UCFSports & J-HMDB & & UCFSports & J-HMDB \\
 \hline
   Linking \cite{gkioxari2015finding} & - & - & & 75.8\% & 53.3\%\\
   
   Affine Tracking \cite{ye2016region} & 90.3\% & 86.6\% & & 74.3\% & 51.7\%\\
   
   Tracking-by-Detection \cite{weinzaepfel2015learning} & 98.7\% & 91.7\% & & 88.2\% & 54.2\%\\

   Tracking-by-Point-Matching & \textbf{99.1\%} & \textbf{92.5\%} & & \textbf{89.6\%} & \textbf{56.3\%}\\
  \hline
\end{tabular}}
\caption{Comparison of different linking and tracking algorithms for action detection on UCFSports and J-HMDB.}
\label{table:track-impact}
\end{table}

\subsubsection{Evaluation of Action Region Detection}
We first compare the static and optical flow images as well as the early- and late-fusion models in Faster R-CNN for the frame-level action region detection. As shown in Table~\ref{table:fusion-impact}, for a single modality, static frames outperform optical flow images. By combining the appearance and motion information in the late-fusion, the performance is boosted from $73.5\%$ to $81.2\%$ on UCFSports and from $54.3\%$ to $56.8\%$ on J-HMDB. By fusing the static and optical flow images as one input and training the network to jointly learn the correlation between the two modalities, the early-fusion improves the detection results by $1.6\%$ and $0.8\%$ on UCFSports and J-HMDB respectively compared against the late-fusion. Moreover, the frame-level detection can be further improved after merging both early- and late-fusion results. Our final action region detection model achieves $84.7\%$ and $59.8\%$ video-mAPs on UCFSports and J-HMDB. This is evident to show the benefits of fusing the appearance and motion information at different stages of Faster R-CNN to provide the complementary and mutually amended action regions.

\begin{table*}
\centering
\resizebox{0.6\linewidth}{!}{
\begin{tabular}{l c c}
 \hline
  Methods & UCFSports & J-HMDB\\ 
 \hline
   Static & 73.5\% & 54.3\%\\
   
   Optical Flow & 72.9\% & 40.6\%\\

   Late Fusion & 81.2\% & 56.8\%\\

   Early Fusion & 82.8\% & 57.6\%\\

   Early + Late Fusions & \textbf{84.7\%} & \textbf{59.8\%}\\
  \hline
\end{tabular}}
\caption{Comparison of different action region detection models by the measurement of frame-mAP.}
\label{table:fusion-impact}
\end{table*}

\subsubsection{Evaluation of Tracking Algorithms} 
We next evaluate the effectiveness of the proposed tracking-by-point-matching algorithm by comparing with the linking method \cite{gkioxari2015finding}, the affine tracking algorithm \cite{ye2016region}, and the traditional tracking-by-detection paradigm \cite{weinzaepfel2015learning} in the context of action detection. In order to make a fair comparison, our tracking-by-point-matching algorithm is performed on the same action region detection method (i.e., R-CNN) as employed in \cite{gkioxari2015finding, weinzaepfel2015learning}. We use the two evaluation metrics recall-track and mAP for the comparison. The recall-track measures how well the generated tubes of an action class cover with the ground-truth tracks, and the mAP reflects the impact of the corresponding tracking method to the final detection results. As shown in Table~\ref{table:track-impact}, our tracking-by-point-matching algorithm outperforms both affine tracking and tracking-by-detection algorithms by the measurement of recall-track. In addition, our approach achieves more significant improvements compared to other methods on UCFSports and J-HMDB in term of both frame-mAP and video-mAP.

\begin{figure}[t]
\centering
\includegraphics[width = 0.8\linewidth]{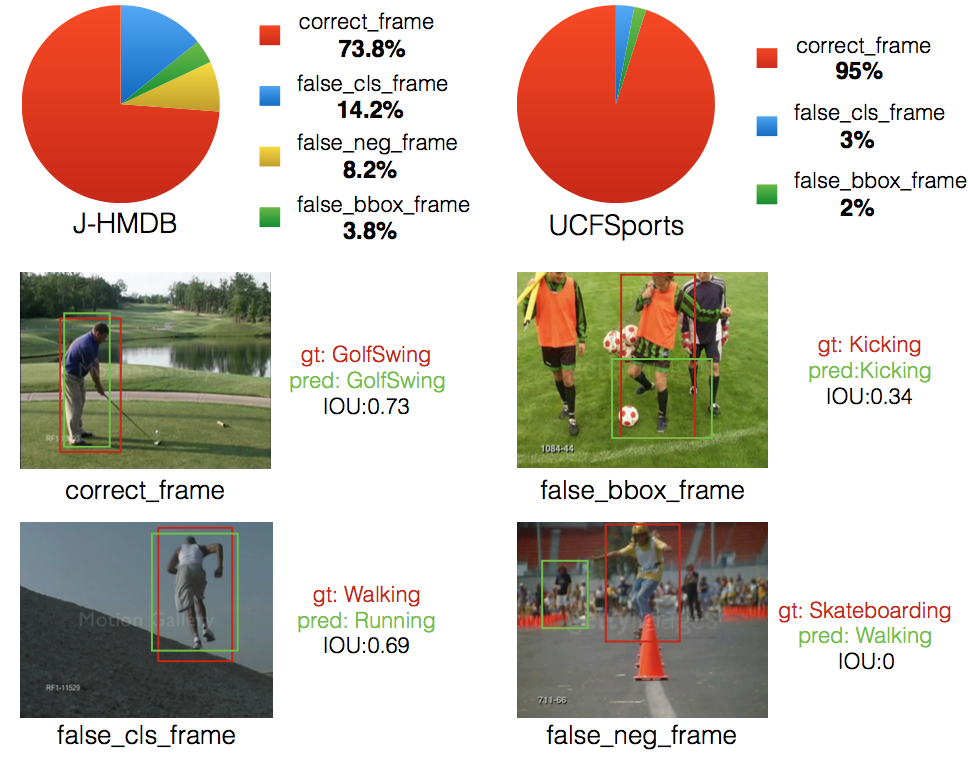}
\caption{Examples of the true and false action detections and the proportion of each detection type in the final results on UCFSports and J-HMDB.}
\label{fig:false}
\end{figure}


\begin{table}[t]
\centering
\resizebox{0.8\linewidth}{!}{
\begin{tabular}{l c c c c c c}
 \hline
  Components & \multicolumn{6}{c}{Combinations}\\ 
 \hline
 \hline
   3DCNN & \cellcolor{gray!25}$\surd$ &\cellcolor{gray!70}$\surd$ & \cellcolor{gray!25}$\surd$& \cellcolor{gray!70}$\surd$&\cellcolor{gray!25}$\surd$&\cellcolor{gray!70}$\surd$\\
   RNN & \cellcolor{gray!25}& \cellcolor{gray!70}$\surd$ &\cellcolor{gray!25} &\cellcolor{gray!70} & \cellcolor{gray!25}&\cellcolor{gray!70}\\
   LSTM & \cellcolor{gray!25}& \cellcolor{gray!70} &\cellcolor{gray!25}$\surd$ &\cellcolor{gray!70} & \cellcolor{gray!25}&\cellcolor{gray!70}\\
   PreRNN & \cellcolor{gray!25}&\cellcolor{gray!70} & \cellcolor{gray!25} &\cellcolor{gray!70}$\surd$ &\cellcolor{gray!25}&\cellcolor{gray!70}$\surd$\\
   Footprint Map & \cellcolor{gray!25}&\cellcolor{gray!70} & \cellcolor{gray!25}& \cellcolor{gray!70} &\cellcolor{gray!25}$\surd$&\cellcolor{gray!70}$\surd$\\
  \hline
  \hline
   Video-mAP & \cellcolor{gray!25}65.6\% & \cellcolor{gray!70}68.7\% & \cellcolor{gray!25}69.1\%&\cellcolor{gray!70}69.4\% &\cellcolor{gray!25}67.5\%&\cellcolor{gray!70}73.8\%\\
  \hline
\end{tabular}}
\caption{Evaluation of different scoring and pruning methods on J-HMDB dataset.}
\label{table:score-impact}
\end{table}

\begin{table}[t]
\centering
\resizebox{0.6\linewidth}{!}{
\begin{tabular}{c c c c}
 \hline
  Methods & UCFSports & J-HMDB & UCF101\\ 
 \hline
   \cite{gkioxari2015finding} & 68.1\% & 36.2\% & -\\
   
   \cite{weinzaepfel2015learning} & 71.9\% & 45.8\% & 35.8\%\\

   \cite{peng2016multi} & 84.5\% & 58.5\% & 65.7\%\\
   
   \cite{Hou2017} & \textbf{86.7\%} & 61.3\% & 41.4\%\\
    
   \cite{Kalogeiton2017} & \textbf{87.7\%} & \textbf{65.7\%} & \textbf{67.1\%}\\

   Ours & \textbf{86.8\%} & \textbf{63.2\%} & \textbf{67.0\%}\\
  \hline
\end{tabular}}
\caption{Comparison to the state-of-the-art methods on  UCFSports, J-HMDB and UCF101 measured by frame-mAP.}
\label{table:frameAP}
\end{table}

\begin{table*}[tp]
\centering
\begin{tabular}{ccc>{\centering}m{0.5cm}c>{\centering}m{0.5cm}c>{\centering}m{0.5cm}c>{\centering}m{0.5cm}cc>{\centering}m{0.5cm}c>{\centering}m{0.5cm}c>{\centering}m{0.5cm}c>{\centering}m{0.5cm}c}
\toprule
\hspace{20pt} & UCFSports &\hspace{15pt} & \multicolumn{4}{c}{J-HMDB} & \hspace{15pt}& \multicolumn{4}{c}{UCF101} \\
\cmidrule{2-2}\cmidrule{4-7}\cmidrule{9-12}
$\sigma$ & 0.5 & \hspace{15pt} & 0.2 & 0.3 & 0.4 & 0.5 & \hspace{15pt} & 0.05 & 0.1 & 0.2 & 0.3 \\
\midrule
\cite{gkioxari2015finding} & 75.8 &  & - & - & - & 53.3 & & - & - & - & - \\
\cite{weinzaepfel2015learning} & 90.5 &  & 63.1 & 63.5 & 62.2 & 60.7 & & 54.3 & 51.7 & 46.8 & 37.8 \\
\cite{online-realtime} & - & & 73.8 & - & - & 72.0 & & - & - & 73.5 & - \\
\cite{peng2016multi} & 94.7 & & 74.3 & - & - & 73.1 & & 78.8 & 77.3 & 72.9 & 65.7 \\
\cite{Kalogeiton2017} & 92.7 & & - & - & - & 73.7 & & - & - & \textbf{77.2} & - \\
Ours & \textbf{95.0} &  & \textbf{75.8} & \textbf{75.2} & \textbf{74.6} & \textbf{73.8} & & \textbf{79.4} & \textbf{77.7} & \textbf{76.2} & \textbf{73.8} \\
\bottomrule
\end{tabular}
\caption{Comparison to the state-of-the-art methods on  UCFSports, J-HMDB and UCF101 measured by video-mAP (\%) under different IOU thresholds.}
\label{table:results}
\end{table*}

\begin{figure*}[t]
\centering
\includegraphics[height=6cm, width = 9cm]{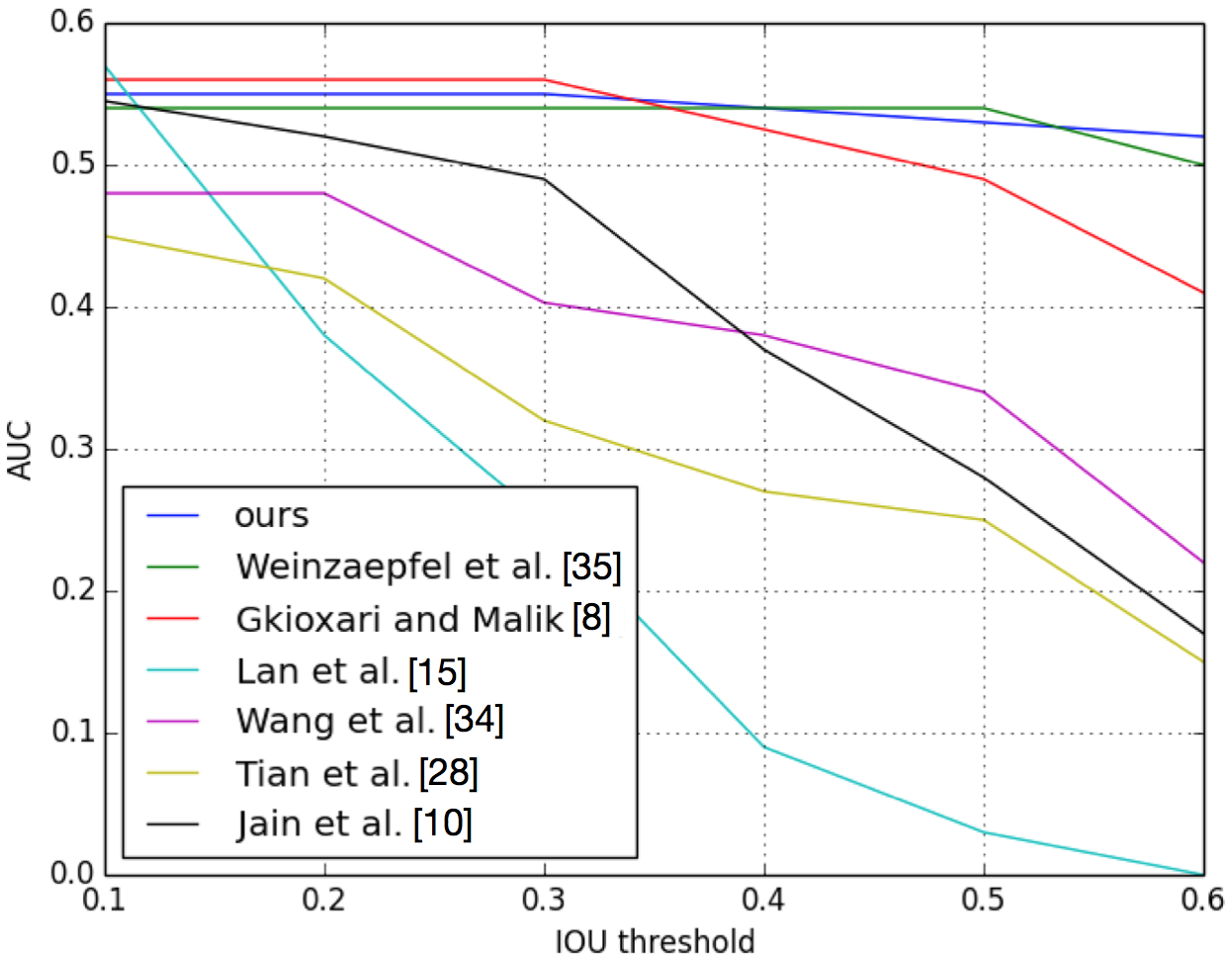}
\caption{AUC values for varying IOU thresholds on UCFSports dataset.}
\label{fig:roc_ucf}
\end{figure*}

\begin{figure*}[t]
\centering
\includegraphics[height=6cm, width = 9cm]{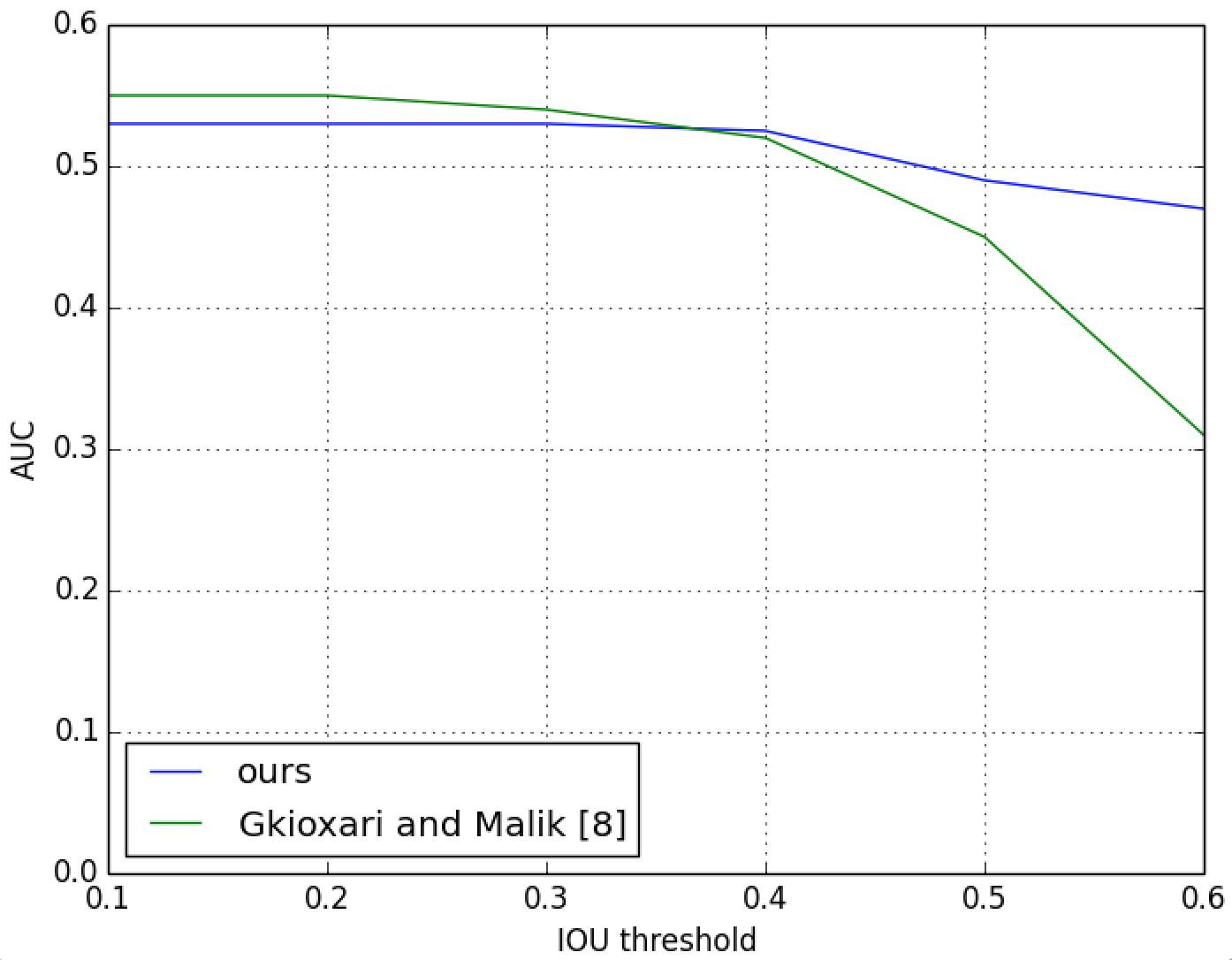}
\caption{AUC values on J-HMDB for a set of IOU thresholds.}
\label{fig:roc_jhmdb}
\end{figure*}

\subsubsection{Evaluation of Action Tube Refinement}
Here we evaluate the impacts of R3DCNN and the action footprint map for refining the candidate tubes. As observed from Table~\ref{table:score-impact}, by modeling the long-term temporal evolution of a tube, 3DCNN with RNN largely improves over 3DCNN which is merely based on individual clips of an action tube that lacks the global temporal context. Furthermore, PreRNN provides an extra boost over the traditional RNN and LSTM. Additionally, our proposed action footprint map consistently improves the results by removing the spatially off-focusing tubes. This clearly shows the advantage of utilizing the action-specific spatial information reserved in the convolutional layers of R3DCNN to clean the false tubes.


\subsubsection{Analysis of False Detections}
We also analyze the constituents of the false instances detected by our method. There are three types of false detections: \textit{false\underline{ }cls\underline{ }frame}, \textit{false\underline{ }bbox\underline{ }frame} and \textit{false\underline{ }neg\underline{ }frame}; where \textit{false\underline{ }cls\underline{ }frame} denotes a detection with a correct spatio-temporal location but assigned a wrong action label; \textit{false\underline{ }bbox\underline{ }frame} indicates a detection with the accurate class label but insufficient IOU ($< 0.5$) with the ground truth; \textit{false\underline{ }neg\underline{ }frame} means that our approach fails to detect a spatio-temporal sequence around the ground truth. Figure~\ref{fig:false} illustrates examples of the three false detections and the proportion of each type in our final results on UCFSports and J-HMDB. From this observation, our method can handle the instances of \textit{false\underline{ }bbox\underline{ }frame} and \textit{false\underline{ }neg\underline{ }frame} quite well, which gives credit to the effectiveness of our tracking-by-point-matching algorithm. The majority of false detections in our approach is \textit{false\underline{ }cls\underline{ }frame}, so exploring more accurate scoring methods to evaluate the credibility of action tubes is one of our future focuses.

\begin{figure*}[t]
\centering
\includegraphics[width = 0.55\linewidth]{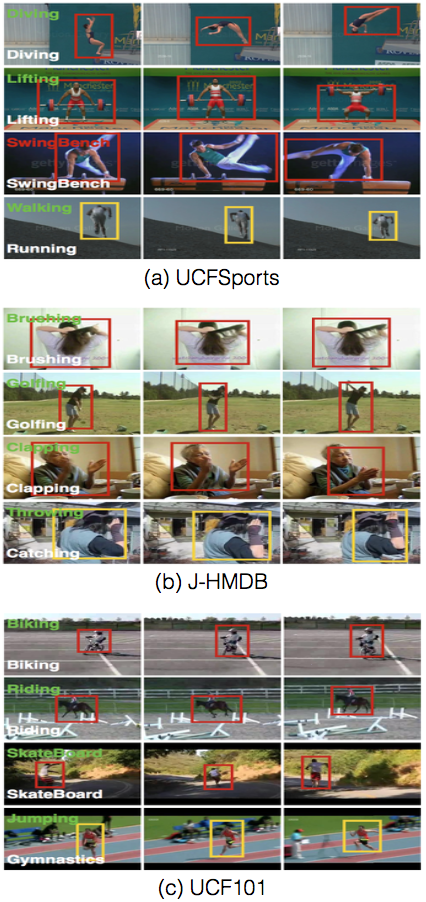}
\caption{Illustration of the action detection results by our method on (a) UCFSports, (b) J-HMDB and (c) UCF101. For each dataset, four videos of different action classes are shown and each of them is sampled by a sequence of three discontinuous frames. In each video sequence, the words in white color are the predicted action classes, and the words in green color denote the ground truth labels. Bounding boxes with red color are the correct detections, and the ones with yellow color are the false detections.}
\label{fig:example}
\end{figure*}

\subsection{Comparison to State-of-the-Art Results}
In this section we compare our approach with the state-of-the-art methods on UCFSports, J-HMDB and UCF101. As observed from Figure~\ref{fig:roc_ucf}, unlike many other methods, whose AUC values vary significantly at different IOU thresholds, our approach performs consistently at both low and high IOU thresholds. As discussed in the paper \cite{Weinzaepfel_2015_ICCV}, this phenomenon demonstrates the high spatial accuracy of our approach, which does not produce many easy negatives even at the low IOU threshold. We also provide the AUC values for a set of IOU thresholds on the J-HMDB dataset in Figure~\ref{fig:roc_jhmdb}.

Both frame-level and video-level mAPs are reported in our experimental results. A detection is correct if its IoU with a ground-truth box or tube is greater than 0.5 and the predicted action label is correct.

\textbf{frame-mAP}. As shown in the Table~\ref{table:frameAP}, our algorithm achieves competitive detection accuracy on the three datasets by the measurement of frame-mAP. Specifically, our method outperforms previous linking based state-of-the-art method \cite{peng2016multi} by $1.9\%$ on UCFSports, $4.7\%$ on J-HMDB and $1.3\%$ on UCF101. This may owe to our proposed tracking algorithm, which exploits the temporal consistency to complement individual detections on each frame.

\textbf{video-mAP}. As shown in Table~\ref{table:results}, our approach achieves superior results over other competing algorithms at various IOU thresholds of UCF101. We note that the paper \cite{peng2016multi} achieves comparable results to our method on UCFSports and J-HMDB datasets, but our result significantly outperforms \cite{peng2016multi} on UCF101 which is a more challenging dataset. We conjecture that UCF101 requires accurate temporal localization and the linking method of \cite{peng2016multi} is not well adapted in finding the precise temporal boundaries. By contrast, our approach performs well in both spatial and temporal localization tasks. Figure~\ref{fig:example} demonstrates some representative examples of successes and failures of the detected actions by our method on the three datasets. As shown in this figure, most of the false detections have close semantics to the ground truth actions, which is also consistent to the analysis of our false detections.


\section{Conclusions}
In this paper, we have proposed an effective framework for spatio-temporal action detection in videos. Specifically, we develop the early- and late-fusion scheme to combine the static and kinematic information through Faster R-CNN to detect action regions at frame-level. A tracking-by-point-matching algorithm is proposed to connect the action regions into action tubes throughout the whole video sequence. Moreover, we employ R3DCNN to classify the tubes and yield accurate temporal boundaries. In the end, we introduce the action footprint map to model the action-specific spatial focus to prune the candidate tubes. In the extensive evaluations, our method achieves superior results on three benchmark datasets.

\section{Acknowledgements}
This work was supported in part by NSF grants EFRI-1137172 and IIS-1400802.

\bibliographystyle{elsarticle-num}
\bibliography{egbib}

\end{document}